\documentclass[10pt,twocolumn,letterpaper]{article}

\usepackage{cvpr}
\usepackage{times}
\usepackage{epsfig}
\usepackage{graphicx}
\usepackage{amsmath}
\usepackage{amssymb}

\usepackage{bbm}
\usepackage{comment}
\usepackage{authblk}

\usepackage[breaklinks=true,bookmarks=false]{hyperref}

\cvprfinalcopy 

\def\cvprPaperID{1535} 

\ifcvprfinal\fi
\begin{document}

\title{Learning Loss for Active Learning}

\author{Donggeun Yoo$^{1,2}$}
\author{In So Kweon$^2$}
\affil{
$^1$Lunit Inc., Seoul, South Korea.\\
$^2$KAIST, Daejeon, South Korea.\\
{\tt\small dgyoo@lunit.io}\quad{\tt\small iskweon77@kaist.ac.kr}
}

\maketitle

\begin{abstract}
The performance of deep neural networks improves with more annotated data. The problem is that the budget for annotation is limited. One solution to this is active learning, where a model asks human to annotate data that it perceived as uncertain. A variety of recent methods have been proposed to apply active learning to deep networks but most of them are either designed specific for their target tasks or computationally inefficient for large networks. In this paper, we propose a novel active learning method that is simple but task-agnostic, and works efficiently with the deep networks. We attach a small parametric module, named ``loss prediction module,'' to a target network, and learn it to predict target losses of unlabeled inputs. Then, this module can suggest data that the target model is likely to produce a wrong prediction. This method is task-agnostic as networks are learned from a single loss regardless of target tasks. We rigorously validate our method through image classification, object detection, and human pose estimation, with the recent network architectures. The results demonstrate that our method consistently outperforms the previous methods over the tasks.
\end{abstract}

\section{Introduction}
\label{sec:introduction}

\begin{figure}[t]
\begin{center}
\small
\includegraphics[width=1\linewidth]{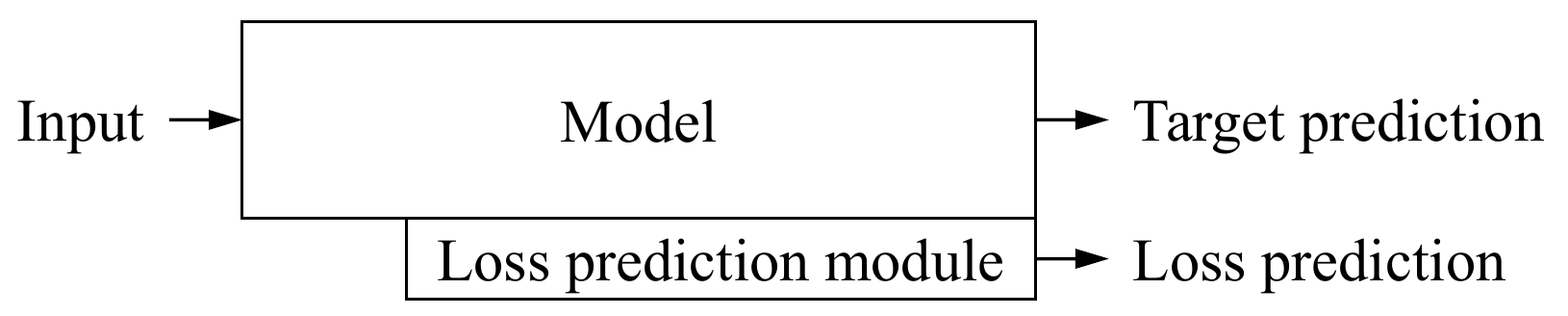}\\
\small(a) A model with a loss prediction module\\
\textcolor{white}{.}\\
\textcolor{white}{.}\\
\includegraphics[width=1\linewidth]{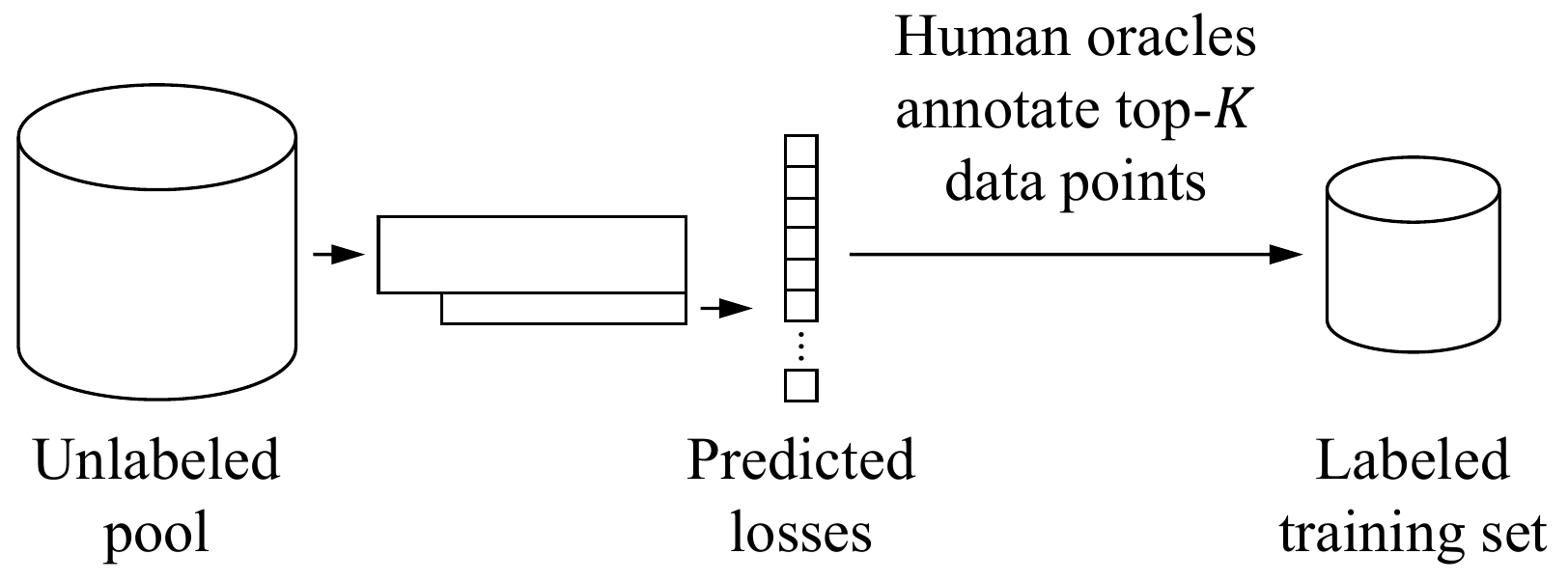}\\
\small(b) Active learning with a loss prediction module\\
\end{center}
\caption{A novel active learning method with a loss prediction module. (a) A loss prediction module attached to a target model predicts the loss value from an input without its label. (b) All data points in an unlabeled pool are evaluated by the loss prediction module. The data points with the top-$K$ predicted losses are labeled and added to a labeled training set.}
\label{fig:overview}
\end{figure}

Data is flooding in, but deep neural networks are still data-hungry. The empirical analysis of \cite{mahajan2018exploring, joulin2016learning} suggests that the performance of recent deep networks is not yet saturated with respect to the size of training data. For this reason, learning methods from semi-supervised learning \cite{rasmus2015semi, papandreou2015weakly, mahajan2018exploring, joulin2016learning} to unsupervised learning \cite{agrawal2015learning, doersch2015unsupervised, zhang2017split, noroozi2017representation} are attracting attention along with weakly-labeled or unlabeled large-scale data.

However, given a fixed amount of data, the performance of the semi-supervised or unsupervised learning is still bound to that of fully-supervised learning. The experimental results of semi-supervised learning in \cite{rasmus2015semi, sener2018active} demonstrate that the higher portion of annotated data ensures superior performance. This is why we are suffering from annotation labor and cost of time.

The cost of annotation varies widely depending on target tasks. In the natural image domain, it is relatively cheap to annotate class labels for classification, but detection requires expensive bounding boxes. For segmentation, it is more expensive to draw pixel-level masks. The situation gets much worse when we consider the bio-medical image domain. It requires board-citified specialists trained for several years (radiologists for radiography images \cite{nam2018development}, pathologists for slide images \cite{lee2018robust}) to obtain annotations.

The budget for annotation is limited. What then is the most efficient use of the budget? \cite{atlas1990training, lewis1994sequential} first proposed active learning where a model actively selects data points that the model is uncertain of. For an example of binary classification~\cite{lewis1994sequential}, the data point whose posterior probability closest to 0.5 is selected, annotated, and added to a training set. The core idea of active learning is that the most informative data point would be more beneficial to model improvement than a randomly chosen data point.

Given a pool of unlabeled data, there have been three major approaches according to the selection criteria: an uncertainty-based approach, a diversity-based approach, and expected model change. The uncertainty approach \cite{lewis1994sequential, joshi2009multi, wang2017cost, tong2001support, seung1992query, beluch2018power} defines and measures the quantity of uncertainty to select uncertain data points, while the diversity approach \cite{sener2018active, nguyen2004active, guo2010active, bilgic2009link} selects diverse data points that represent the whole distribution of the unlabeled pool. Expected model change \cite{roy2001toward, settles2008multiple, freytag2014selecting} selects data points that would cause the greatest change to the current model parameters or outputs if we knew their labels. Readers can review most of classical studies for these approaches in \cite{settles2012active}.

The simplest method of the uncertainty approach is to utilize class posterior probabilities to define uncertainty. The probability of a predicted class \cite{lewis1994sequential} or an entropy of class posterior probabilities \cite{joshi2009multi, wang2017cost} defines uncertainty of a data point. Despite its simplicity, this approach has performed remarkably well in various scenarios. For more complex recognition tasks, it is required to re-define task-specific uncertainty such as object detection \cite{wang2018towards}, semantic segmentation \cite{liu2017active}, and human pose estimation \cite{dutt2016active}.

As a task-agnostic uncertainty approach, \cite{seung1992query, beluch2018power} train multiple models to construct a committee, and measure the consensus between the multiple predictions from the committee. However, constructing a committee is too expensive for current deep networks learned with large data. Recently, Gal \textit{et al.}~\cite{gal2017deep} obtains uncertainty estimates from deep networks through multiple forward passes by Monte Carlo Dropout~\cite{gal2016dropout}. It was shown to be effective for classification with small datasets, but according to \cite{sener2018active}, it does not scale to larger datasets.

The distribution approach could be task-agnostic as it depends on a feature space, not on predictions. However, extra engineering would be necessary to design a location-invariant feature space for localization tasks such as object detection and segmentation. The method of expected model change has been successful for small models but it is computationally impractical for recent deep networks.

The majority of empirical results from previous researches suggest that active learning is actually reducing the annotation cost. The problem is that most of methods require task-specific design or are not efficient in the recent deep networks, resulting in another engineering cost. In this paper, we aim to propose a novel active learning method that is simple but task-agnostic, and performs well on deep networks.

A deep network is learned by minimizing a single loss, regardless of what a task is, how many tasks there are, and how complex an architecture is. This fact motivates our task-agnostic design for active learning. If we can predict the loss of a data point, it becomes possible to select data points that are expected to have high losses. The selected data points would be more informative to the current model.

To realize this scenario, we attach a ``\textit{loss prediction module}'' to a deep network and learn the module to predict the loss of an input data point. The module is illustrated in Figure~\ref{fig:overview}-(a). Once the module is learned, it can be utilized to active learning as shown in Figure~\ref{fig:overview}-(b). We can apply this method to any task that uses a deep network.

We validate the proposed method through image classification, human pose estimation, and object detection. The human pose estimation is a typical regression task, and the object detection is a more complex problem combined with both regression and classification. The experimental results demonstrate that the proposed method consistently outperforms previous methods with a current network architecture for each recognition task. To the best of our knowledge, this is the first work verified with three different recognition tasks using the state-of-the-art deep network models.

\subsection{Contributions}
In summary, our major contributions are
\begin{enumerate}
    \item Proposing a simple but efficient active learning method with the loss prediction module, which is directly applicable to any tasks with recent deep networks.
    \item Evaluating the proposed method with three learning tasks including classification, regression, and a hybrid of them, by using current network architectures.
\end{enumerate}

\section{Related Research}
\label{sec:related_research}
Active learning has advanced for more than a couple of decades. First, we introduce classical active learning methods that use small-scale models \cite{settles2012active}. In the uncertainty approach, a naive way to define uncertainty is to use the posterior probability of a predicted class \cite{lewis1994sequential, lewis1994heterogeneous}, or the margin between posterior probabilities of a predicted class and the secondly predicted class \cite{joshi2009multi, roth2006margin}. The entropy \cite{settles2008analysis, luo2013latent, joshi2009multi} of class posterior probabilities generalizes the former definitions. For SVMs, distances \cite{tong2001support, vijayanarasimhan2014large, li2014multi} to the decision boundaries can be used to define uncertainty. Another approach is the query-by-committee \cite{seung1992query, mccallumzy1998employing, iglesias2011combining}. This method constructs a committee comprising multiple independent models, and measures disagreement among them to define uncertainty.

The distribution approach chooses data points that represent the distribution of an unlabeled pool. The intuition is that learning over a representative subset would be competitive over the whole pool. To do so, \cite{nguyen2004active} applies a clustering algorithm to the pool, and \cite{yang2015multi, elhamifar2013convex, guo2010active} formulate the subset selection as a discrete optimization problem. \cite{bilgic2009link, hasan2015context, mac2014hierarchical} consider how close a data point is to surrounding data points to choose one that could well propagate the knowledge. The method of expected model change is a more sophisticated and decision-theoretic approach for model improvement. It utilizes the current model to estimate expected gradient length \cite{settles2008multiple}, expected future errors \cite{roy2001toward}, or expected output changes \cite{freytag2014selecting, kading2016active}, to all possible labels.

Do these methods, advanced with small models and data, well scale to large deep networks \cite{krizhevsky2012imagenet, he2016deep} and data? Fortunately, the uncertainty approach \cite{lin2018active, wang2017cost} for classification tasks still performs well despite its simplicity. However, a task-specific design is necessary for other tasks since it utilizes network outputs. As a more generalized uncertainty approach, \cite{gal2017deep} obtains uncertainty estimates through multiple forward passes with Monte Carlo Dropout, but it is computationally inefficient for recent large-scale learning as it requires dense dropout layers that drastically slow down the convergence speed. This method has been verified only with small-scale classification tasks. \cite{beluch2018power} constructs a committee comprising 5 deep networks to measure disagreement as uncertainty. It has shown the state-of-the-art classification performance, but it is also inefficient in terms of memory and computation for large-scale problems.

Sener \textit{et al.}~\cite{sener2018active} propose a distribution approach on an intermediate feature space of a deep network. This method is directly applicable to any task and network architecture since it depends on intermediate features rather than the task-specific outputs. However, it is still questionable whether the intermediate feature representation is effective for localization tasks such as detection and segmentation. This method has also been verified only with classification tasks. As the two approaches based on uncertainty and distribution are differently motivated, they are complementary to each other. Thus, a variety of hybrid strategies have been proposed \cite{liu2017active, zhou2017fine, paul2017non, yang2017suggestive} for their specific tasks. 

Our method can be categorized into the uncertainty approach but differs in that it predicts ``\textit{loss}'' based on the input contents, rather than statistically estimating uncertainty from outputs. It is similar to a variety of hard example mining \cite{shrivastava2016training, felzenszwalb2010object} since they regard training data points with high losses as being significant for model improvement. However, ours is distinct from theirs in that we do not have annotations of data. 

\section{Method}
\label{sec:method}
In this section, we introduce the proposed active learning method. We start with an overview of the whole active learning system in Section~\ref{sec:overview}, and provide in-depth descriptions of the loss prediction module in Section~\ref{sec:loss_prediction_module}, and the method to learn this module in Section~\ref{sec:learning_to_learn_loss}.

\subsection{Overview}
\label{sec:overview}
In this section, we formally define the active learning scenario with the proposed loss prediction module. In this scenario, we have a set of models composed of a target model $\Theta_\text{target}$ and a loss prediction module $\Theta_\text{loss}$. The loss prediction module is attached to the target model as illustrated in Figure~\ref{fig:overview}-(a). The target model conducts the target task as $\hat{y}=\Theta_\text{target}(x)$, while the loss prediction module predicts the loss $\hat{l}=\Theta_\text{loss}(h)$. Here, $h$ is a feature set of $x$ extracted from several hidden layers of $\Theta_\text{target}$.

In most real-world learning problems, we can gather a large pool of unlabeled data $\mathcal{U}_N$ at once. The subscript $N$ denotes the number of data points. Then, we uniformly sample $K$ data points at random from the unlabeled pool, and ask human oracles to annotate them to construct an initial labeled dataset $\mathcal{L}^0_K$. The subscript $0$ means it is the initial stage. This process reduces the size of the unlabeled pool as $\mathcal{U}^0_{N-K}$.

Once the initially labeled dataset $\mathcal{L}_K^0$ is obtained, we jointly learn an initial target model $\Theta^0_\text{target}$ and an initial loss prediction module $\Theta_\text{loss}^0$. After initial training, we evaluate all the data points in the unlabeled pool by the loss prediction module to obtain data-loss pairs $\{(x,\hat{l})|x\in \mathcal{U}_{N-K}^0\}$. Then, human oracles annotate the data points of the $K$-highest losses. The labeled dataset $\mathcal{L}_K^0$ is updated with them and becomes $\mathcal{L}_{2K}^1$. After that, we learn the model set over $\mathcal{L}_{2K}^1$ to obtain  $\{\Theta^1_\text{target}, \Theta^1_\text{loss}\}$. This cycle, illustrated in Figure~\ref{fig:overview}-(b), repeats until we meet a satisfactory performance or until we have exhausted the budget for annotation.

\subsection{Loss Prediction Module}
\label{sec:loss_prediction_module}
The loss prediction module is core to our task-agnostic active learning since it learns to imitate the loss defined in the target model. This section describes how we design it.

The loss prediction module aims to minimize the engineering cost of defining task-specific uncertainty for active learning. Moreover, we also want to minimize the computational cost of learning the loss prediction module, as we are already suffering from the computational cost of learning very deep networks. To this end, we design a loss prediction module that is (1) much smaller than the target model, and (2) jointly learned with the target model. There is no separated stage to learn this module.

\begin{figure}[t]
\begin{center}
\includegraphics[width=1\linewidth]{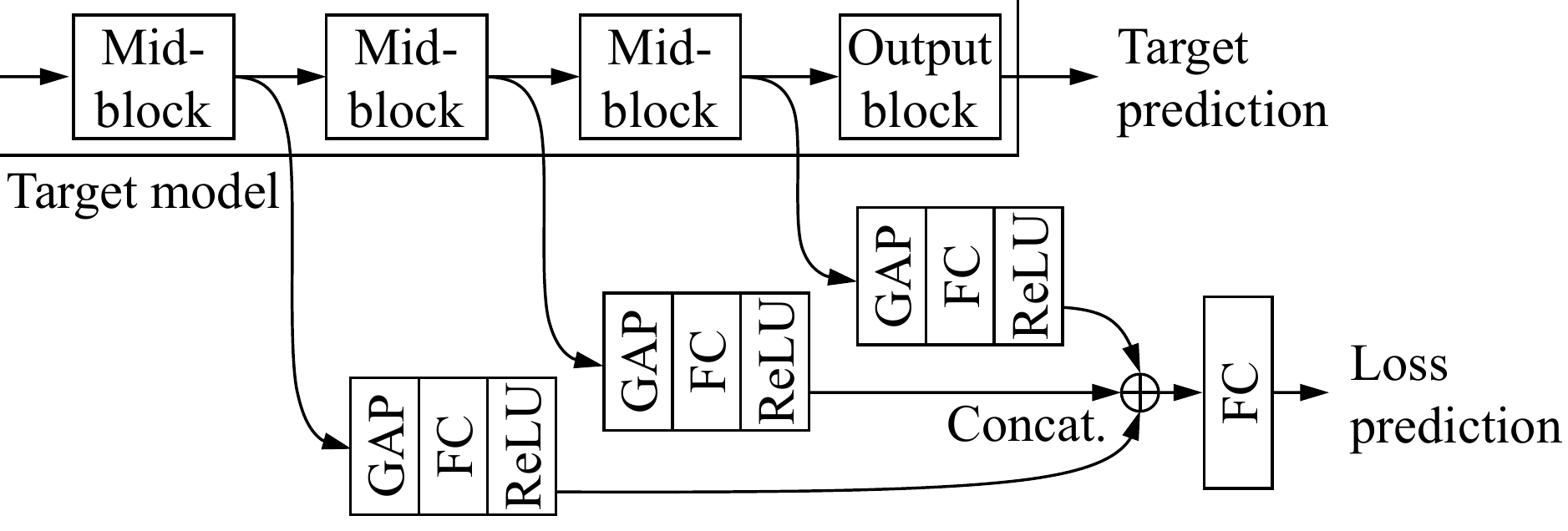}\\
\end{center}
\caption{The architecture of the loss prediction module. This module is connected to several layers of the target model to take multi-level knowledge into consideration for loss prediction. The multi-level features are fused and map to a scalar value as the loss prediction.}
\label{fig:architecture}
\end{figure}

Figure~\ref{fig:architecture} illustrates the architecture of our loss prediction module. It takes multi-layer feature maps $h$ as inputs that are extracted between the mid-level blocks of the target model. These multiple connections let the loss prediction module to choose necessary information between layers useful for loss prediction. Each feature map is reduced to a fixed dimensional feature vector through a global average pooling (GAP) layer and a fully-connected layer. Then, all features are concatenated and pass through another fully-connected layer, resulting in a scalar value $\hat{l}$ as a predicted loss. Learning this two-story module requires much less memory and computation than the target model. We have tried to make this module deeper and wider, but the performance does not change much.
 
\subsection{Learning Loss}
\label{sec:learning_to_learn_loss}
In this section, we provide an in-detail description of how to learn the loss prediction module defined before. Let us suppose we start the $s$-th active learning stage. We have a labeled dataset $\mathcal{L}^s_{K\cdot (s+1)}$ and a model set composed of a target model $\Theta_\text{target}$ and a loss prediction module $\Theta_\text{loss}$. Our objective is to learn the model set for this stage $s$ to obtain $\{\Theta^s_\text{target}, \Theta^s_\text{loss}\}$.

\begin{figure}[t]
\begin{center}
\includegraphics[width=1\linewidth]{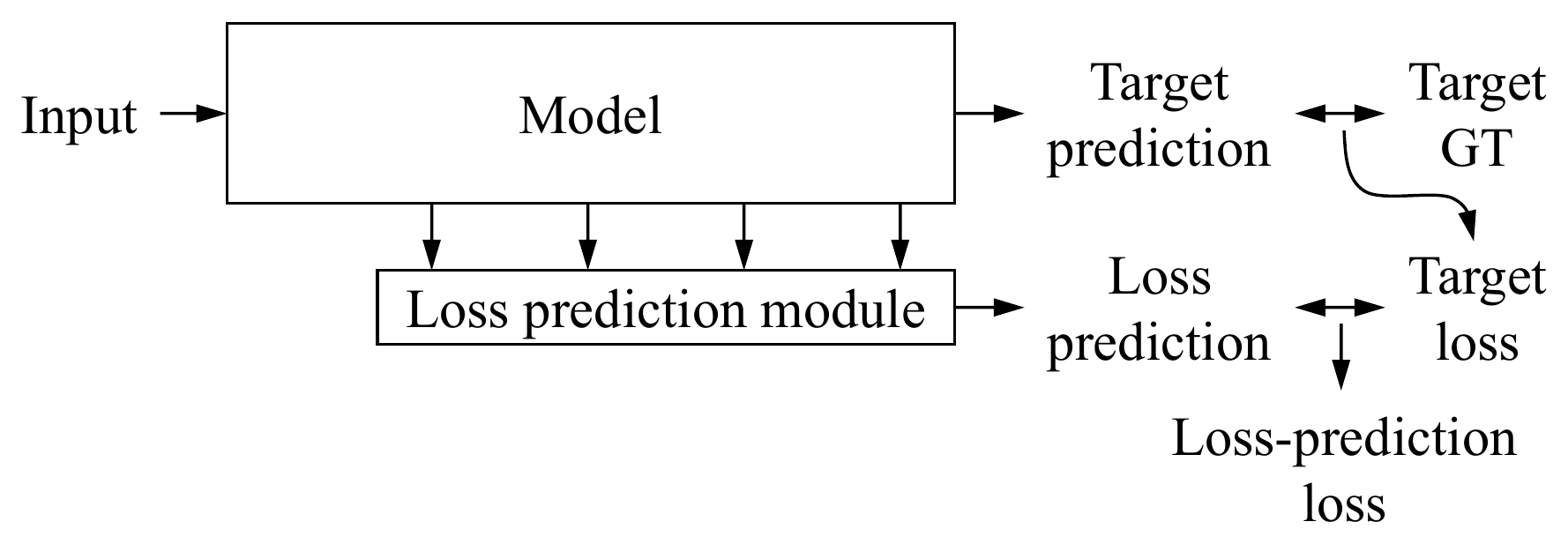}\\
\end{center}
\caption{Method to learn the loss. Given an input, the target model outputs a target prediction, and the loss prediction module outputs a predicted loss. The target prediction and the target annotation are used to compute a target loss to learn the target model. Then, the target loss is regarded as a ground-truth loss for the loss prediction module, and used to compute the loss-prediction loss.}
\label{fig:learning}
\end{figure}

Given a training data point $x$, we obtain a target prediction through the target model as $\hat{y}=\Theta_\text{target}(x)$, and also a predicted loss through the loss prediction module as $\hat{l}=\Theta_\text{loss}(h)$. With the target annotation $y$ of $x$, the target loss can be computed as $l=L_\text{target}(\hat{y}, y)$ to learn the target model. Since this loss $l$ is a ground-truth target of $h$ for the loss prediction module, we can also compute the loss for the loss prediction module as $L_\text{loss}(\hat{l}, l)$. Then, the final loss function to jointly learn both of the target model and the loss prediction module is defined as
\begin{equation}
\label{eq:final_loss1}
L_\text{target}(\hat{y},y)+\lambda\cdot L_\text{loss}(\hat{l}, l)
\end{equation}
where $\lambda$ is a scaling constant. This procedure to define the final loss is illustrated in Figure~\ref{fig:learning}. 

Perhaps the simplest way to define the loss-prediction loss function is the mean square error (MSE) $L_\text{loss}(\hat{l}, l)=(\hat{l}-l)^2$. However, MSE is not a suitable choice for this problem since the scale of the real loss $l$ changes (decreases in overall) as learning of the target model progresses. Minimizing MSE would let the loss prediction module adapt roughly to the scale changes of the loss $l$, rather than fitting to the exact value. We have tried to minimize MSE but failed to learn a good loss prediction module, and active learning with this module actually demonstrates performance worse than previous methods.

It is necessary for the loss-prediction loss function to discard the overall scale of $l$. Our solution is to compare a pair of samples. Let us consider a training iteration with a mini-batch $\mathcal{B}^s\subset \mathcal{L}^s_{K\cdot (s+1)}$. In the mini-batch whose size is $B$, we can make $B/2$ data pairs such as $\{x^p=(x_i,x_j)\}$. The subscript $p$ represents that it is a pair, and the mini-batch size $B$ should be an even number. Then, we can learn the loss prediction module by considering the difference between a pair of loss predictions, which completely make the loss prediction module discard the overall scale changes. To this end, the loss function for the loss prediction module is defined as
\begin{multline}
\label{eq:loss_prediction_loss}
L_\text{loss}(\hat{l^p}, l^p) = \max\left(0, -\mathbbm{1}(l_i,l_j)\cdot(\hat{l_i}-\hat{l_j})+\xi\right)\\
\text{s.t.}\quad\mathbbm{1}(l_i,l_j)=
\left\{\begin{aligned}
+1,&\quad\textrm{if}\quad l_i>l_j\\
-1,&\quad\textrm{otherwise}
\end{aligned}\right.
\end{multline}
where $\xi$ is a pre-defined positive margin and the subscript $p$ also represents the pair of $(i,j)$. For instance when $l_i>l_j$, this function states that no loss is given to the module only if $\hat{l_i}$ is larger than $\hat{l_j}+\xi$, but otherwise a loss is given to the module to force it to increase $\hat{l_i}$ and decrease $\hat{l_j}$.

Given a mini-batch $\mathcal{B}^s$ in the active learning stage $s$, our final loss function to jointly learn the target model and the loss prediction module is
\begin{multline}
\label{eq:final_loss2}
\frac{1}{B}\sum_{(x,y)\in\mathcal{B}^s}L_\text{target}(\hat{y},y)+\lambda\frac{2}{B}\cdot \sum_{(x^p,y^p)\in\mathcal{B}^s}L_\text{loss}(\hat{l^p}, l^p)\\
\text{s.t.}\quad
\begin{aligned}
&\hat{y}=\Theta_\text{target}(x) \\
&\hat{l^p}=\Theta_\text{loss}(h^p) \\
&l^p=L_\text{target}(\hat{y^p},y^p).
\end{aligned}
\end{multline}

Minimizing this final loss give us $\Theta^s_\text{loss}$ as well as $\Theta^s_\text{target}$ without any separated learning procedure nor any task-specific assumption. The learning process is efficient as the loss prediction module $\Theta^s_\text{loss}$ has been designed to contain a small number of parameters but to utilize rich mid-level representations $h$ of the target model. This loss prediction module will pick the most informative data points and ask human oracles to annotate them for the next active learning stage $s+1$.

\section{Evaluation}
\label{sec:evaluation}
In this section, we rigorously evaluate our method through three visual recognition tasks. To verify whether our method works efficiently regardless of tasks, we choose diverse target tasks including image classification as a classification task, object detection as a hybrid task of classification and regression, and human pose estimation as a typical regression problem. These three tasks are indeed important research topics for visual recognition in computer vision, and are very useful for many real-world applications.

We have implemented our method and all the recognition tasks with PyTorch~\cite{paszke2017automatic}. For all tasks, we initialize a labeled dataset $\mathcal{L}_K^0$ by randomly sampling $K$=1,000 data points from the entire dataset $\mathcal{U}_N$. In each active learning cycle, we continue to train the current model by adding $K$=1,000 labeled data points. The margin $\xi$ defined in the loss function (Equation~\ref{eq:loss_prediction_loss}) is set to 1. We design the fully-connected layers (FCs) in Figure~\ref{fig:architecture} except for the last one to produce a 128-dimensional feature. For each active learning method, we repeat the same experiment multiple times with different initial labeled datasets, and report the performance mean and standard deviation. For each trial, our method and compared methods share the same random seed for a fair comparison. Other implementation details, datasets, and experimental results for each task are described in the following Sections~\ref{sec:image_classification}, \ref{sec:object_detection}, \ref{sec:human_pose_estimation}.

\subsection{Image Classification}
\label{sec:image_classification}
Image classification is a common problem that has been verified by most of the previous active learning methods. In this problem, a target model recognizes the category of a major object from an input image, so object category labels are required for supervised learning.

\paragraph{Dataset}
We choose CIFAR-10 dataset~\cite{krizhevsky2009learning} as it has been used for recent active learning methods \cite{sener2018active,beluch2018power}. CIFAR-10 consists of 60,000 images of 32$\times$32$\times$3 size, assigned with one of 10 object categories. The training and test sets contain 50,000 and 10,000 images respectively. We regard the training set as the initial unlabeled pool $\mathcal{U}_\text{50,000}$. As studied in \cite{sener2018active,settles2012active}, selecting $K$-most uncertain samples from such a large pool $\mathcal{U}_\text{50,000}$ often does not work well, because image contents among the $K$ samples are overlapped. To address this, \cite{beluch2018power} obtains a random subset $\mathcal{S}_M\subset \mathcal{U}_N$ for each active learning stage and choose $K$-most uncertain samples from $\mathcal{S}_M$. We adopt this simple yet efficient scheme and set the subset size to $M$=10,000. As an evaluation metric, we use the classification accuracy.

\paragraph{Target model}
We employ the 18-layer residual network (ResNet-18) \cite{he2016deep} as we aim to verify our method with current deep architectures. We have utilized an open source\footnote{https://github.com/kuangliu/pytorch-cifar} in which this model specified for CIFAR showing 93.02\% accuracy is implemented. ResNet-18 for CIFAR is identical to the original ResNet-18 except for the first convolution and pooling layers. The first convolution layer is changed to contain 3$\times$3 kernels with the stride of 1 and the padding of 1, and the max pooling layer is dropped, to adapt to the small size images of CIFAR.

\paragraph{Loss prediction module}
ResNet-18 is composed of 4 basic blocks $\{\text{conv}i\_1,\text{conv}i\_2\;|\;i\text{=}2,3,4,5\}$ following the first convolution layer. Each block comprises two convolution layers. We simply connect the loss prediction module to each of the basic blocks to utilize the 4 rich features from the blocks for estimating the loss.

\paragraph{Learning}
For training, we apply a standard augmentation scheme including 32$\times$32 size random crop from 36$\times$36 zero-padded images and random horizontal flip, and normalize images using the channel mean and standard deviation vectors estimated over the training set. For each of active learning cycle, we learn the model set $\{\Theta^s_\text{target}, \Theta^s_\text{loss}\}$ for 200 epochs with the mini-batch size of 128 and the initial learning rate of 0.1. After 160 epochs, we decrease the learning rate to 0.01. The momentum and the weight decay are 0.9 and 0.0005 respectively. After 120 epochs, we stop the gradient from the loss prediction module propagated to the target model. We set $\lambda$ that scales the loss-prediction loss in Equation~\ref{eq:final_loss2} to 1.

\paragraph{Comparison targets}
We compare our method with random sampling, entropy-based sampling \cite{settles2008analysis,luo2013latent}, and core-set sampling \cite{sener2018active}, which is a recent distribution approach. For the entropy-based method, we compute the entropy from a softmax output vector. For core-set, we have implemented $K$-Center-Greedy algorithm in~\cite{sener2018active} since it is simple to implement yet marginally worse than the mixed integer program. We also run the algorithm over the last feature space right before the classification layer as \cite{sener2018active} do. Note that we use \textit{exactly the same hyper-parameters} to train target models for all methods including ours.

\begin{figure}[t]
\begin{center}
\includegraphics[width=0.85\linewidth]{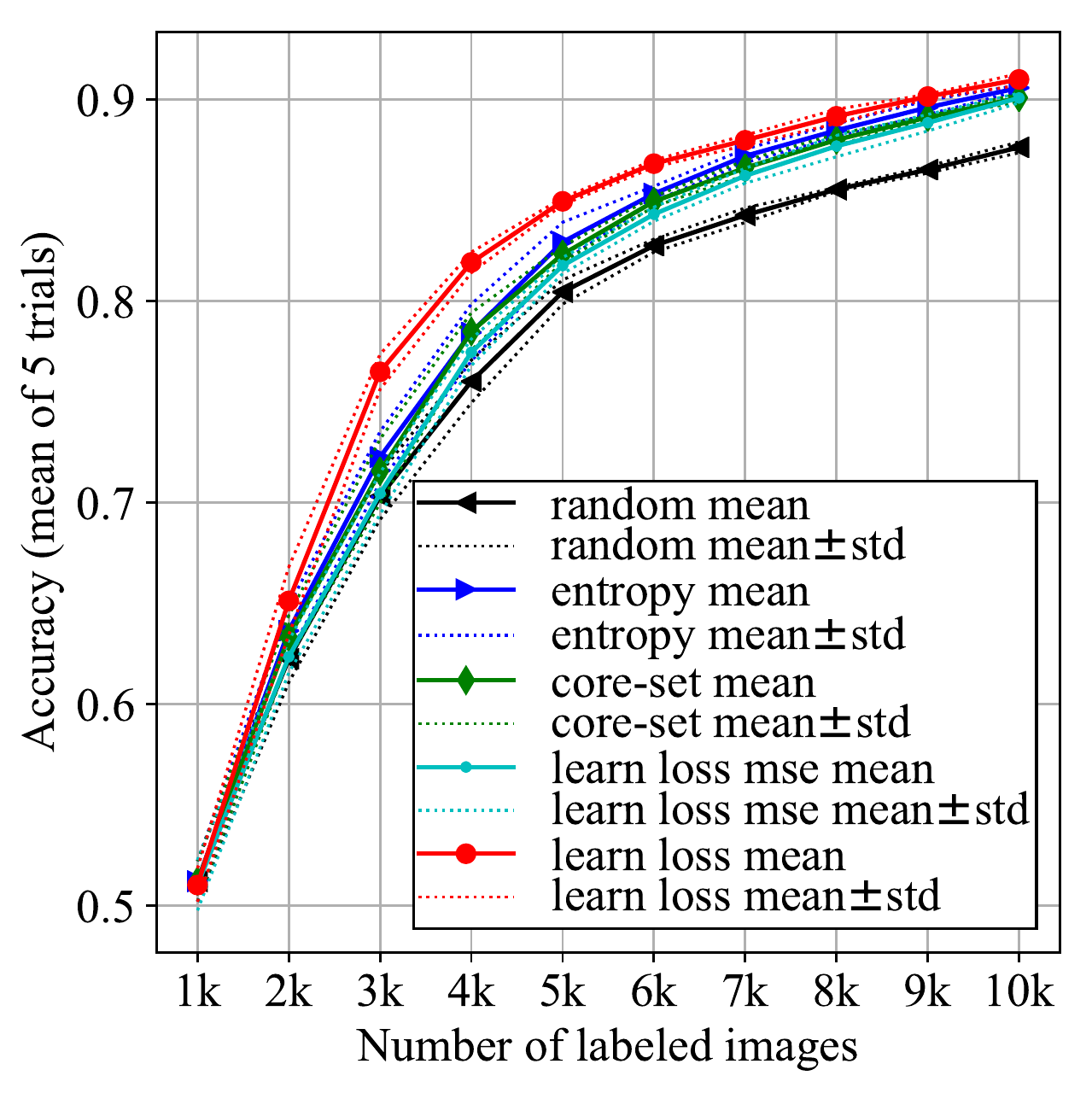}
\caption{Active learning results of image classification over CIFAR-10.}
\label{fig:cifar_comparison}
\end{center}
\end{figure}

The results are shown in Figure~\ref{fig:cifar_comparison}. Each point is an average of 5 trials with different initial labeled datasets. Our implementations show that both entropy-based and core-set methods have better results than the random baseline. In the last active learning cycle, the entropy and core-set methods show 0.9059 and 0.9010 respectively, while the random baseline shows 0.8764. The performance gaps between these methods are similar to those of \cite{beluch2018power}. In particular, the simple entropy-based method works very effectively with the classification which is typically learned to minimize cross-entropy between predictions and target labels.

Our method noted as ``learn loss'' shows the highest performance for all active learning cycles. In the last cycle, our method achieves an accuracy of 0.9101. This is 0.42\% higher than the entropy method and 0.91\% higher than the core-set method. Although the performance gap to the entropy-based method is marginal in classification, our method can be effectively applied to more complex and diverse target tasks.

\begin{figure}[t]
\begin{center}
\includegraphics[width=0.87\linewidth]{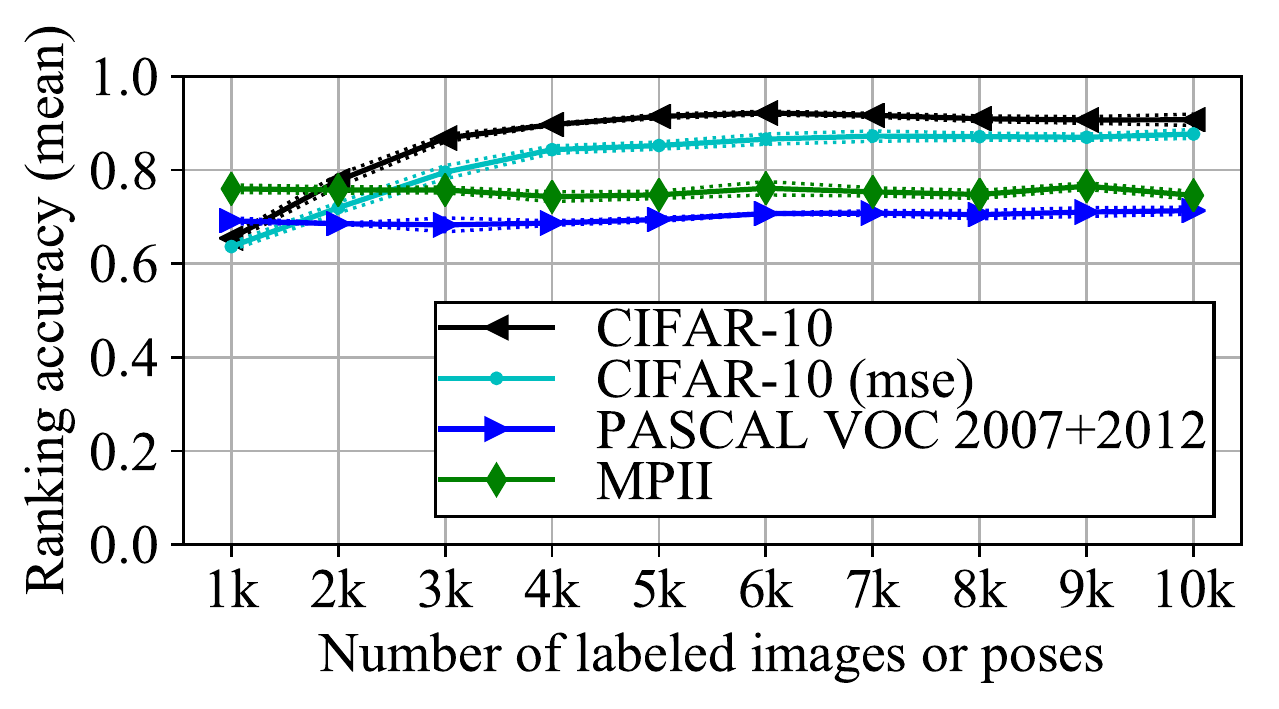}
\caption{Loss-prediction accuracy of the loss prediction module.}
\label{fig:ranking_accuracy}
\end{center}
\end{figure}

We define an evaluation metric to measure the performance of the loss prediction module. For a pair of data points, we give a score 1 if the predicted ranking is true, and 0 for otherwise. These binary scores from every pair of test sets are averaged to a value named ``ranking accuracy''. Figure~\ref{fig:ranking_accuracy} shows the ranking accuracy of the loss prediction module over the test set. As we add more labeled data, loss prediction module becomes more accurate and finally reaches 0.9074. The use of MSE for learning the loss prediction module with $\lambda$=0.1, noted by ``learn loss mse'', yields lower loss-prediction performance (Figure~\ref{fig:ranking_accuracy}) that results in less-efficient active learning (Figure~\ref{fig:cifar_comparison}).

\subsection{Object Detection}
\label{sec:object_detection}
Object detection localizes bounding boxes of semantic objects and recognizes the categories of the objects. It is a typical hybrid task as it combines a regression problem for bounding box estimation and a classification problem for category recognition. It requires both object bounding boxes and category labels for supervised learning.

\paragraph{Dataset}
We evaluate our method on PASCAL VOC 2007 and 2012 \cite{Everingham10} that provide full bounding boxes of 20 object categories. VOC 2007 comprises \texttt{\small trainval'07} and \texttt{\small test'07} which contain 5,011 images and 4,952 images respectively. VOC 2012 provides 11,540 images as \texttt{\small trainval'12}. Following the recent use of VOC for object detection, we make a super-set \texttt{\small trainval'07+12} by combining the two, and use it as the initial unlabeled pool $\mathcal{U}_\text{16,551}$. The active learning method is evaluated over \texttt{\small test'07} with mean average precision (mAP), which is a standard metric for object detection. We do not create a random subset $\mathcal{S}_M$ since the size of the pool $\mathcal{U}_\text{16,551}$ is not very large in contrast to CIFAR-10.

\paragraph{Target model}
We employ Single Shot Multibox Detector (SSD) \cite{liu2016ssd} as it is one of the popular models for recent object detection. It is a large network with a backbone of VGG-16~\cite{simonyan2014very}. We have utilized an open source\footnote{https://github.com/amdegroot/ssd.pytorch} which shows 0.7743 (mAP) slightly higher than the original paper.

\paragraph{Loss prediction module}
SSD estimates bounding-boxes and their classes from 6-level feature maps extracted from $\{\text{conv}i\;|\;i\text{=}4\_3,7,8\_2,9\_2,10\_2,11\_2\}$~\cite{liu2016ssd}. Accordingly, we also connect the loss prediction module to each of them to utilize the 6 rich features for estimating the loss.

\paragraph{Learning}
We use exactly the same hyper-parameter values and the data augmentation scheme described in \cite{liu2016ssd}, except for the number of iterations since we use a smaller training set for each active learning cycle. We learn the model set for 300 epochs with the mini-batch size of 32. After 240 epochs, we reduce the learning rate from 0.001 to 0.0001. We set the scaling constant $\lambda$ in Equation~\ref{eq:final_loss2} to 1.

\paragraph{Comparison targets}
For the entropy-based method, we compute the entropy of an image by averaging all entropy values from softmax outputs corresponding to detection boxes. For core-set, we also run $K$-Center-Greedy over conv7 (i.e., FC7 in VGG-16) features after applying the spatial average pooling. Note, we use exactly the same hyper-parameters to train SSDs for all methods including ours.

\begin{figure}[t]
\begin{center}
\includegraphics[width=0.87\linewidth]{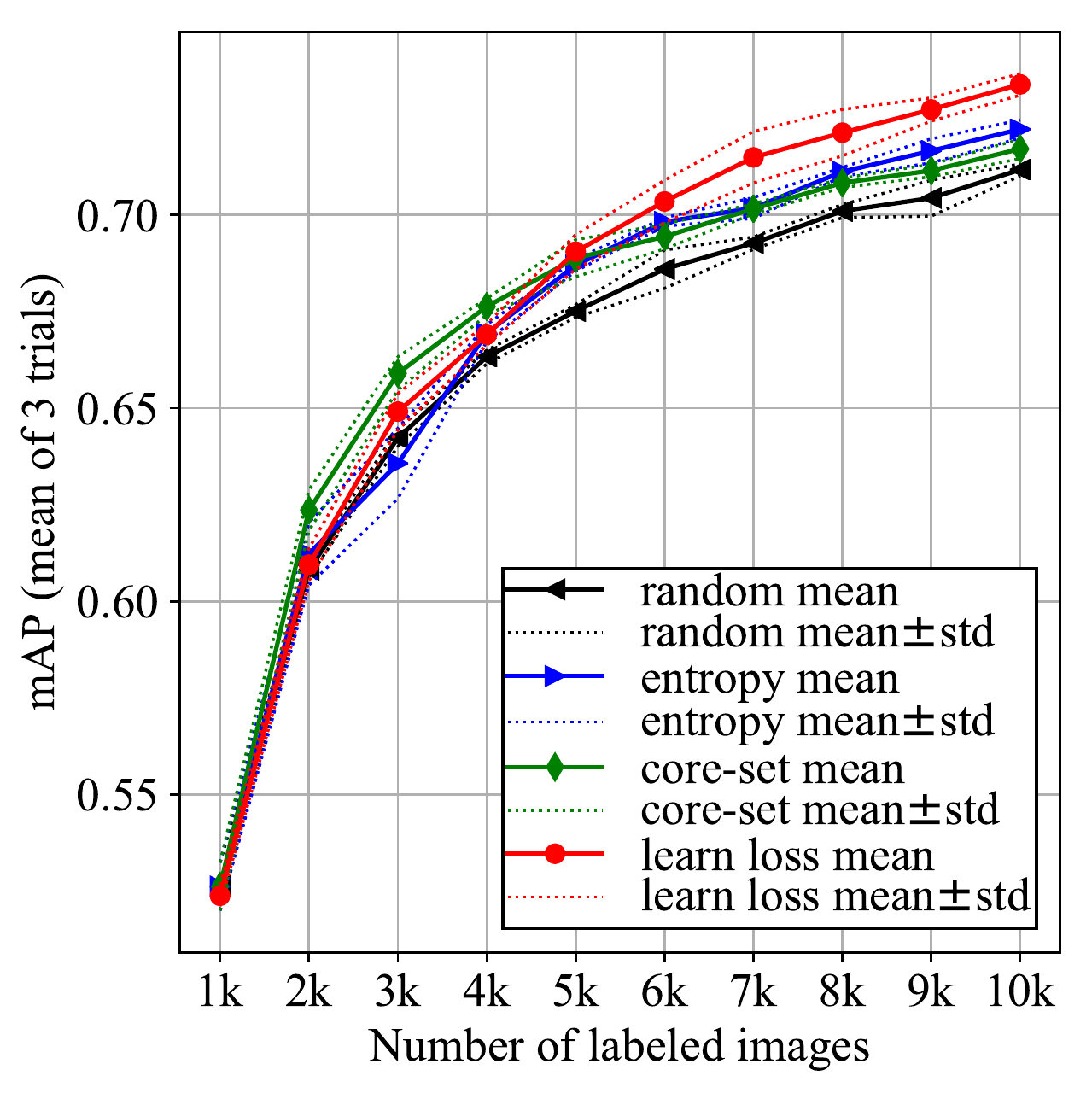}
\caption{Active learning results of object detection over PASCAL VOC 2007+2012.}
\label{fig:voc_comparison}
\end{center}
\end{figure}

Figure~\ref{fig:voc_comparison} shows the results. Each point is an average of 3 trials with different initial labeled datasets. In the last active learning cycle, our method achieves 0.7338 mAP which is 2.21\% higher than 0.7117 of the random baseline. The entropy and core-set methods, showing 0.7222 and 0.7171 respectively, also perform better than the random baseline. However, our method outperforms these methods by margins of 1.15\% and 1.63\%. The entropy method cannot capture the uncertainty about bounding box regression, which is an important element of object detection, so need to design another uncertainty metric specified for regression. The core-set method also needs to design a feature space that well encodes object-centric information while being invariant to object locations. In contrast, our learning-based approach does not need specific designs since it predicts the final loss value, regardless of tasks. Even if it is much difficult to predict the final loss come from regression and classification, our loss prediction module yields about 70\% ranking accuracy as shown in Figure~\ref{fig:ranking_accuracy}.

\subsection{Human Pose Estimation}
\label{sec:human_pose_estimation}
Human pose estimation is to localize all the body parts from an image. The point annotations of all the body parts are required for supervised learning. It is often approached by a regression problem as the target is a set of points.

\paragraph{Dataset}
We choose MPII dataset~\cite{andriluka20142d} which is commonly used for the majority of recent works. We follow the same splits used in~\cite{newell2016stacked} where a training set consists of 22,246 poses from 14,679 images and a test set consists of 2,958 poses from 2,729 images. We use the training set as the initial unlabeled pool $\mathcal{U}_\text{22,246}$. For each cycle, we obtain a random sub-pool $\mathcal{S}_{5,000}$ from $\mathcal{U}_{22,246}$, following the similar portion of the sub-pool to the entire pool in CIFAR-10. The standard evaluation metric for this problem is Percentage of Correct Key-points (PCK) which measures the percentage of predicted key-points falling within a distance threshold to the ground truth. Following~\cite{newell2016stacked}, we use PCKh@0.5 in which the distance is normalized by a fraction of the head size and the threshold is 0.5.

\paragraph{Target model}
We adopt Stacked Hourglass Networks~\cite{newell2016stacked}, in which an hourglass network consists of down-scale pooling and subsequent up-sampling processes to allow bottom-up, top-down inference across scales. The network produces heatmaps corresponding to the body parts and they are compared to ground-truth heatmaps by applying an MSE loss. We have utilized an open source\footnote{https://github.com/bearpaw/pytorch-pose} yielding 88.78\% (PCK@0.5), which is similar to \cite{newell2016stacked} with 8 hourglass networks. Since learning 8 hourglass networks on a single GPU with the original mini-batch size of 6 is too slow for our active learning experiments, we have tried multi-GPU learning with larger mini-batch sizes. However, the performance has significantly decreased as the mini-batch size increases, even without the loss prediction module. Thus, we have inevitably stacked two hourglass networks which show 86.95\%.

\paragraph{Loss prediction module}
For each hourglass network, the body part heatmaps are driven from the last feature map of (H,W,C)=(64,64,256). We choose this feature map to estimate the loss. As we stack two hourglass networks, the two feature maps are given to our loss prediction module.

\paragraph{Learning}
We use exactly the same hyper-parameter values and data augmentation scheme described in \cite{newell2016stacked}, except the number of training iterations. We learn the model set for 125 epochs with the mini-batch size of 6. After 100 epochs, we reduce the learning rate from 0.00025 to 0.000025. After 75 epochs, the gradient from the loss prediction module is not propagated to the target model. We set the scaling constant $\lambda$ in Equation~\ref{eq:final_loss2} to 0.0001 since the scale of MSE is very small (around 0.001 after several epochs).

\paragraph{Comparison targets}
Stacked Hourglass Networks do not produce softmax outputs but body part heatmaps. Thus, we apply the softmax to each heatmap and estimate an entropy for each body part. We then average all of the entropy values. For core-set, we run $K$-Center-Greedy over the last feature maps after applying the spatial average pooling. Note, we use exactly the same hyper-parameters to train the target models for all methods including ours.

\begin{figure}[t]
\begin{center}
\includegraphics[width=0.85\linewidth]{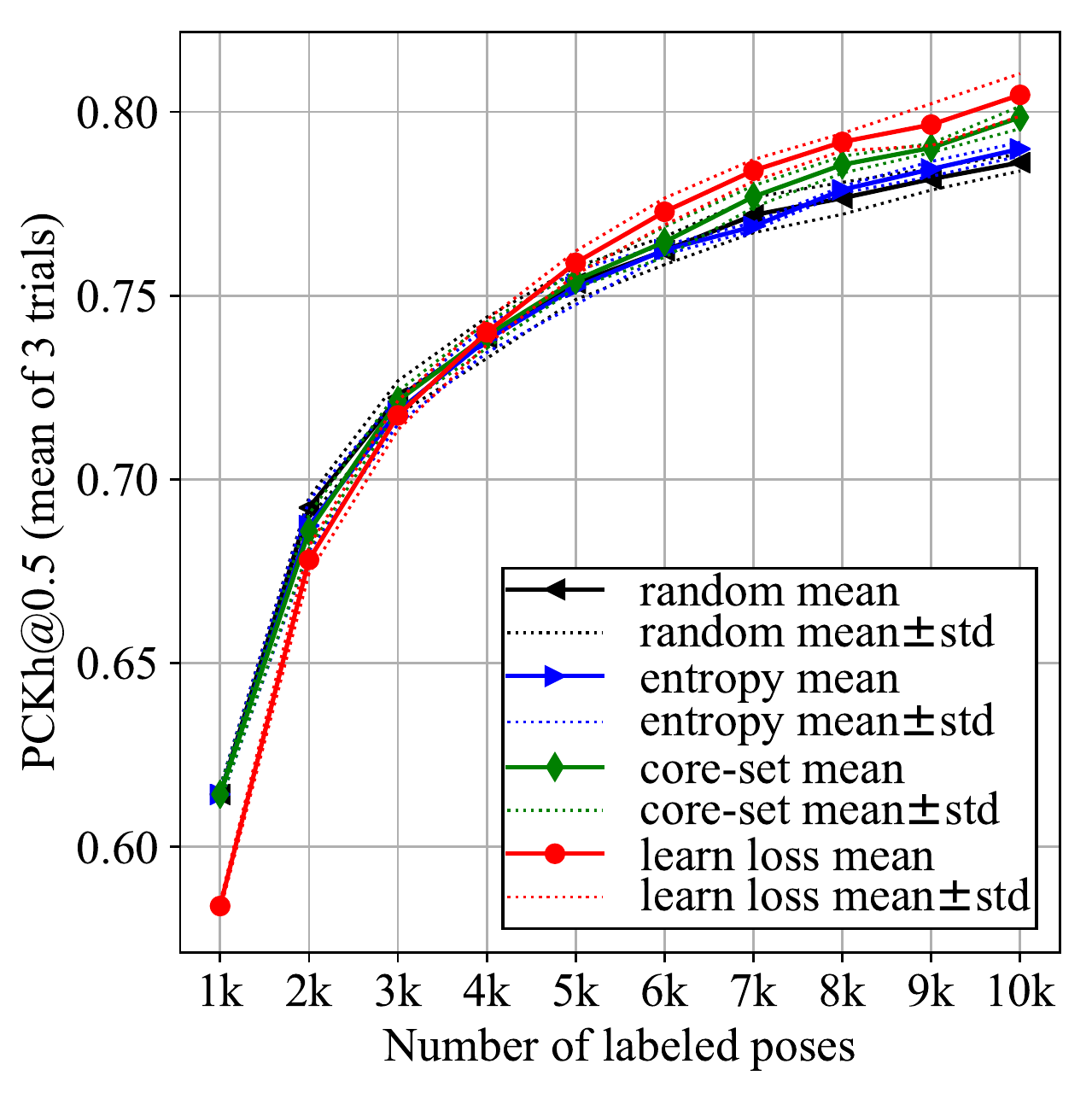}
\caption{Active learning results of human pose estimation over MPII.}
\label{fig:mpii_comparison}
\end{center}
\end{figure}

Experiment results are given in Figure~\ref{fig:mpii_comparison}. Each point is also an average of 3 trials with different initial labeled datasets. The results show that our method outperforms other methods as the active learning cycle progresses. At the end of the cycles, our method attains 0.8046 PCKh@0.5 while the entropy and core-set methods reach 0.7899 and 0.7985, respectively. The performance gaps to these methods are 1.47\% and 0.61\%. The random baseline shows the lowest of 0.7862. In human pose estimation, the entropy method is not as effective as the classification problem. While this method is advantageous to classification in which a cross-entropy loss is directly minimized, this task minimizes an MSE to estimate body part heatmaps. The core-set method also requires a novel feature space that is invariant to the body part location while preserving the local body part features.

\begin{figure}[t]
\begin{center}
\includegraphics[width=0.85\linewidth]{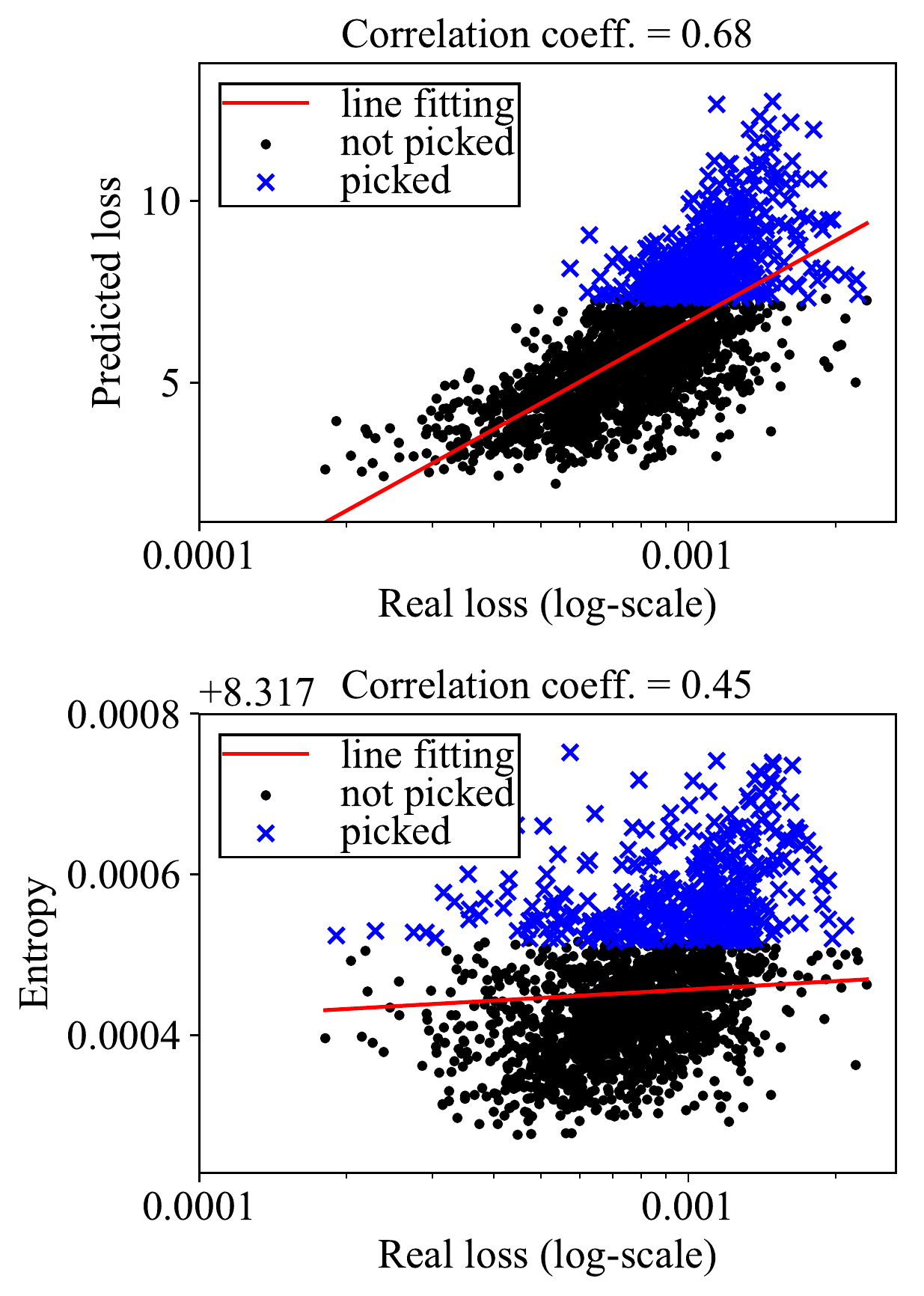}
\caption{Data visualization of (top) our method and (bottom) entropy-based method. We use the model set from the last active learning cycle to obtain the loss, predicted loss and entropy of a human pose. 2,000 poses randomly chosen from the MPII test set are shown.}
\label{fig:loss_pairs}
\end{center}
\end{figure}

Our loss prediction module predicts the regression loss with about 75\% of ranking accuracy (Figure~\ref{fig:ranking_accuracy}), which enables efficient active learning in this problem. We visualize how the predicted loss correlates with the real loss in Figure~\ref{fig:loss_pairs}. At the top of the figure, the data points of the MPII test set are scattered to the axes of predicted loss and real loss. Overall, the two values are correlated, and the correlation coefficient~\cite{boddy2009statistical} (0 for no relation, 1 for strong relation) is 0.68. At the bottom of the figure, the data points are scattered to the axes of entropy and real loss. The correlation coefficient is 0.45, which is much lower than our predicted loss. The blue color means 20\% data points selected from the population according to the predicted loss or entropy. The points chosen by our method actually have high loss values, while the entropy method chooses many points with low loss values. This visualization demonstrates that our method is effective for selecting informative data points.

\section{Limitations and Future Work}
\label{sec:limitations_and_future_work}
We have introduced a novel active learning method that is applicable to current deep networks with a wide range of tasks. The method has been verified with three major visual recognition tasks with popular network architectures. Although the uncertainty score provided by this method has been effective, the diversity or density of data was not considered. Also, the loss prediction accuracy was relatively low in complex tasks such as object detection and human pose estimation. We will continue this research to take data distribution into consideration and design a better architecture and objective function to increase the accuracy of the loss prediction module.

{\small
\bibliographystyle{ieee}
\bibliography{egbib}
}

\end{document}